\documentclass[twoside,11pt]{article}
\usepackage{todonotes}
\presetkeys{todonotes}{inline}{}

\usepackage{jmlr2e}

\jmlrheading{1}{2021}{1-48}{4/00}{10/00}{meila00a}{Shengyi Huang, Rousslan Fernand Julien Dossa, and Chang Ye, Jeff Braga}

\ShortHeadings{CleanRL}{Shengyi, Rousslan, Chang, Jeff}
\firstpageno{1}

\begin{document}

\title{CleanRL: High-quality Single-file Implementations of Deep Reinforcement Learning Algorithms}

\author{\name Shengyi Huang$^{1}$ \email costa.huang@outlook.com\\
       \name Rousslan Fernand Julien Dossa$^{2}$ \email doss@ai.cs.kobe-u.ac.jp\\
       \name Chang Ye$^{3}$ \email c.ye@nyu.edu \\
       \name Jeff Braga$^{1}$ \email \email jeffreybraga@gmail.com  \\
    %   \name Santiago Onta\~{n}\'{o}n$^{1} \\
       \addr $^{1}$College of Computing and Informatics, Drexel University, USA\\
       \addr $^{2}$Graduate School of System Informatics, Kobe University, Japan\\
       \addr $^{3}$Tandon School of Engineering, New York University, USA }

\editor{PLACEHOLDER}

\maketitle

\begin{abstract}%   <- trailing '%' for backward compatibility of .sty file
\texttt{CleanRL} is an open-source library that provides high-quality single-file implementations of Deep Reinforcement Learning algorithms. It provides a simpler yet scalable developing experience by having a straightforward codebase and integrating production tools to help interact and scale experiments. In CleanRL, we put all details of an algorithm into a single file,  making these performance-relevant details  easier to recognize. Additionally, an experiment tracking feature is available to help log metrics, hyperparameters, videos of an agent's gameplay, dependencies, and more to the cloud. Despite succinct implementations, we have also designed tools to help scale, at one point orchestrating experiments on more than 2000 machines simultaneously via Docker and cloud providers. Finally,  we have ensured the quality of the implementations by benchmarking against a variety of environments.
% \texttt{CleanRL} is an open-source library that provides high-quality single-file implementations of Deep Reinforcement Learning (DRL) algorithms. \texttt{CleanRL} provides an easier yet scalable developing experience by having a simpler code base and integrating with production tools to help interact and scale experiments. By putting all the relevant implementation details of an algorithm into a single file, \texttt{CleanRL}'s code is easier to read. Additionally, an experiment tracking feature is available to help log the metrics, hyperparameters, videos of an agent's gameplay, dependencies, and more to the cloud. 
% Despite the simplistic implementation, we have ensured the quality of the implementation by benchmarking against a variety of environments.
% Finally, we have also designed tools to help scale, at one point orchestrating more than 2000 concurrent experiments at the same time via Docker and cloud providers.

% with research-friendly features. The implementation is clean and simple, yet we can scale it to run thousands of experiments using AWS Batch. 
\end{abstract}

% \begin{keywords}
%   Bayesian Networks, Mixture Models, Chow-Liu Trees
% \end{keywords}

\section{Introduction}
% \texttt{rllab}~\citep{rllab} and its successor, \texttt{garage}, provide systematic benchmarking of continuous-action algorithms on their own benchmark environments.
% \texttt{Dopamine}~\citep{dopamine} primarily focuses on DQN and its extensions for discrete-action environments.
% \texttt{rlpyt}~\citep{rlpyt} supports both discrete and continuous-action algorithms from the three classes: policy gradient (with V-functions), deep Q-learning, and policy gradient with Q-functions.
% Other libraries also support diverse sets of algorithms~\citep{baselines, coach, stablebaselines, rayrllib}.
% \texttt{catalyst.RL}~\citep{catalyst_rl}

In recent years, Deep Reinforcement Learning (DRL) algorithms have achieved great success in training autonomous agents for tasks ranging from playing video games directly from pixels to robotic control \citep{mnih2013playing,lillicrap2015continuous,schulman2017proximal}. At the same time, open-source DRL libraries also flourish in the community, such as \texttt{Stable Baselines 3 (SB3)}~\citep{stable-baselines3}, \texttt{RLLib}~\citep{liang2018rllib}, \texttt{MushroomRL}~\citep{d2020mushroomrl}, \texttt{PFRL}~\citep{fujita2021chainerrl}, and others. Many of them have adopted good modular designs and fostered vibrant development communities. Nevertheless, understanding all implementation details of an algorithm remains difficult  because these details are spread to different modules. However, it has become increasingly important to understand these details because implementations matter~\citep{engstrom2019implementation}.

In this paper, we introduce \texttt{CleanRL}, a DRL library based on single-file implementations to help researchers understand all algorithms' details, prototype new features, analyze experiments, and scale with ease. \texttt{CleanRL} is a \emph{non-modular} library. Each algorithm in \texttt{CleanRL} is self-contained in a single file. We have trimmed the lines of code (LOC) to a minimal degree while maintaining readability and  decent performance w.r.t reputable reference implementations. For instance, our Proximal Policy Optimization (PPO)~\citep{schulman2017proximal} implementation matches \texttt{SB3}'s PPO performance in the Atari game Breakout using only 337 LOC in a single file\footnote{\url{https://github.com/vwxyzjn/cleanrl/blob/2486b90102bd87fa43cda481fe711e0eda0c162b/cleanrl/ppo_atari.py}}, making it much easier to understand the algorithm in one go. In contrast, the researchers of other libraries need to understand the modularity design (on average 7-15 files) which can be comprised of thousands of LOC.

% In comparison, most other libraries would implement using 1000+ LOC in multiple files, which makes for a more general setup but would inevitably make it harder to understand the algorithm in whole.

\begin{figure*}[t]
    \centering
    % \makebox[\textwidth][c]{\includegraphics[width=1.2\textwidth]{images/W&B.pdf}}%
    \includegraphics[width=1\columnwidth]{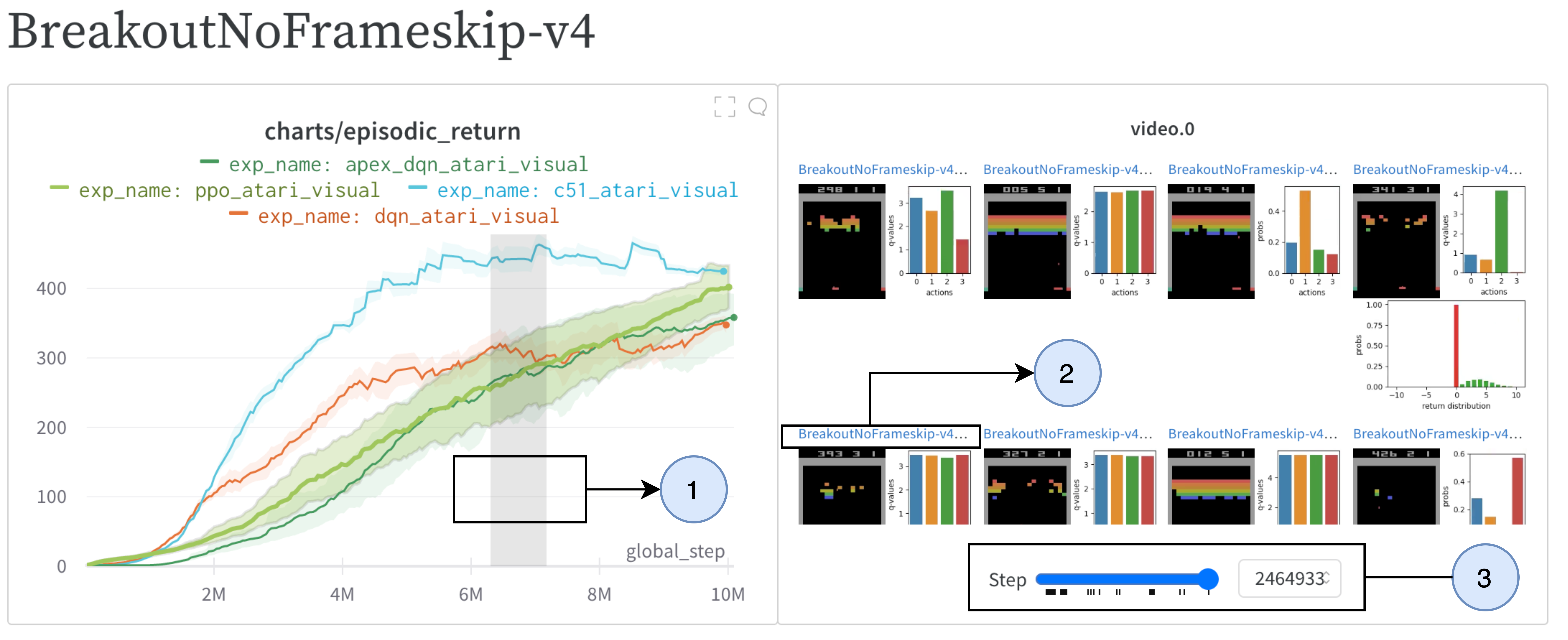}
    \caption{A screenshot of the Open RL Benchmark where the users can \textbf{(1)} zoom into the learning curve, \textbf{(2)} click to trace back to the original experiment for more info (e.g. hyperparameters), and \textbf{(3)} checkout videos at different training stages.}
    \label{fig:wandb}
\end{figure*}
% Although this design would result in duplicate code, we are able to implement the algorithms with significantly fewer lines-of-code, making them almost look like pseudocode. For example, we have implemented Proximal Policy Optimization with 11 core implementation details using 321 LOC.

% able to implement Proximal 
% This obviously result in duplicate code in common setups 
% and most of our algorithm implementation only depends on \texttt{PyTorch}, \texttt{NumPy}, \texttt{Gym}, with no other inter-library reference.

% While \texttt{CleanRL} is stand-alone, for convenience, it supports optional integration with  \texttt{Weights and Biases (W\&B, \url{https://wandb.ai})} for experiment tracking and \texttt{AWS Batch} for experiment orchestration. Notably, our architecture design allows the users to flexibly swap them out, using different tracking and orchestration vendors. 

While \texttt{CleanRL} is stand-alone, for convenience, it supports optional integration with production-quality tool providers. Firstly, integrating \texttt{Weights and Biases (W\&B)} helps track the logs and metrics to the cloud. Over the years, we have tracked thousands of experiments across 7+ algorithms and 40+ games in our Open RL Benchmark (\url{https://benchmark.cleanrl.dev}) to ensure and showcase the quality of our implementations (see Figure~\ref{fig:wandb}). Secondly, integrating \texttt{AWS Batch} helps provision any AWS instances as computational nodes for large-scale experiment orchestration.  Adopting these tools has helped \texttt{CleanRL} focus just on algorithms without worrying about analysis or scaling utilities. Also, our architecture design allows the users to flexibly use other tracking and orchestration vendors if desired, so there is no dependency on proprietary services.

\section{Single-file Implementations}
% Even small implementation details can hugely impact the performance of deep RL algorithms~\citep{engstrom2019implementation}, and it has become increasingly important to understand them. \texttt{CleanRL} helps the researchers to recognize these details by adopting single-file implementations. That is, we put the \textbf{algorithm details} and the \textbf{environment-specific details} into a single file. 
Even small implementation details can hugely impact the performance of deep RL algorithms~\citep{engstrom2019implementation},  however the nature of the modular libraries encapsulates and hides theses details. \texttt{CleanRL} helps the researchers to recognize these details by adopting single-file implementations. That is, we put the \textbf{algorithm details} and the \textbf{environment-specific details} into a single file. 
% each algorithm with environment-specific details is being coded into a single file. 
For example, we have a 
\begin{enumerate}
    \item \texttt{ppo.py} (321 LOC) for the classic control environments, such as \texttt{CartPole-v1}.
    \item \texttt{ppo\_atari.py} (337 LOC) for the Atari environments~\citep{bellemare13arcade}.
    \item \texttt{ppo\_continuous\_action.py} (331 LOC) for the robotics environments (e.g. MuJoCo, PyBullet) with continuous action spaces~\citep{schulman2017proximal}.
\end{enumerate}
Despite having duplicate code among these files, the single-file implementations have the following benefits.
\begin{enumerate}
    \item \textbf{Transparent learning experience} It becomes easier to recognize all aspects of the code in one place. By looking at \texttt{ppo.py}, it is straightforward to recognize the 11 general implementation details of PPO, especially when watching the corresponding tutorial\footnote{See \url{https://youtu.be/MEt6rrxH8W4}}. Then, comparing it with \texttt{ppo\_atari.py} shows about 30 LOC difference to implement 9 Atari-specific details\footnote{See \url{https://youtu.be/05RMTj-2K_Y}}, and another comparison with \texttt{ppo\_continuous\_action.py} shows  about 25 LOC difference to implement 6 details for continuous action spaces\footnote{See \url{https://youtu.be/_rDUEOsjUds}}. 
    \item \textbf{Faster debug experience} Most variables in the files live in the \emph{global python name scope}. This means the researchers can do \texttt{Ctrl+C} to stop the program execution and check most variables and their shapes in the interactive shell (see Appendix~A). This is arguably more convenient than using the Python's debugger, which only shows the variables in a specific name scope like in a function.
    \item \textbf{Faster prototype experience} Because of the faster debug experience, it becomes easier to develop new features. For example, invalid action masking~\citep{huang2020closer} is a common technique used in games with large parameterized action spaces. With \texttt{CleanRL}, it takes about 40 LOC to implement\footnote{\url{https://www.diffchecker.com/g9TYkKCS}}. In comparison, it takes substantially more LOC (e.g. 600+ LOC not counting test cases) because of overhead such as re-factoring the functional arguments, making the classes more general.
\end{enumerate}
% \noindent \textbf{Easier learning experience} 
% helps researchers understand all aspects of the code and makes it easier to prototype.
% In this example, \texttt{ppo.py} showcases the 11 core implementation details
For this motivation, we have also implemented \texttt{dqn.py}, \texttt{dqn\_atari.py}~\citep{mnih2013playing}, \texttt{c51.py}, \texttt{c51\_atari.py}~\citep{bellemare2017distributional}, \texttt{apex\_atari.py}~\citep{horgan2018distributed},  \texttt{ddpg\_continuous\_action.py}~\citep{lillicrap2015continuous}, \texttt{td3\_continuous\_action.py}~\citep{fujimoto2018addressing}, and \texttt{sac\_continuous\_action.py}~\citep{haarnoja2018soft}.

% \todo{Global name scope}

% contains the 11 general implementation and works with 

\section{Experiment Tracking and Analysis}
\texttt{CleanRL} supports integration with \texttt{W\&B}, a modern experiment tracking and analysis solution.  When running an experiment, \texttt{W\&B} automatically tracks the 1) source code, 2) dependency (i.e. \texttt{requirements.txt﻿}), 3) hyperparameters, 4) the exact terminal command used to run the experiment, 5) training metrics, 6) videos of the agents playing the game, 7) system metrics, and 8) logs (\texttt{stdout, stderr}). See Appendix B for 9 selected features that help query and clean the experiment data.

Although \texttt{W\&B} is a proprietary service that we elect to use, our implementation uses \texttt{Tensorboard} to record metrics, which is open-source. So the users could easily change about 10 LOC to use other experiment tracking vendors such as \texttt{ClearML (\url{https://clear.ml})} that supports \texttt{Tensorboard} integration.

\begin{figure*}[t]
\centering
{\includegraphics[width=0.7\textwidth]{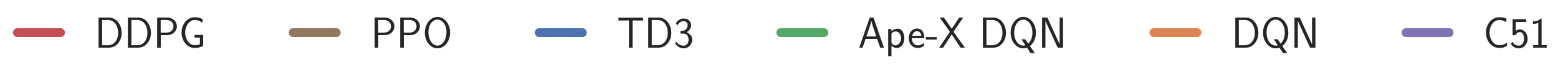}}\hfill
{\includegraphics[width=0.32\textwidth]{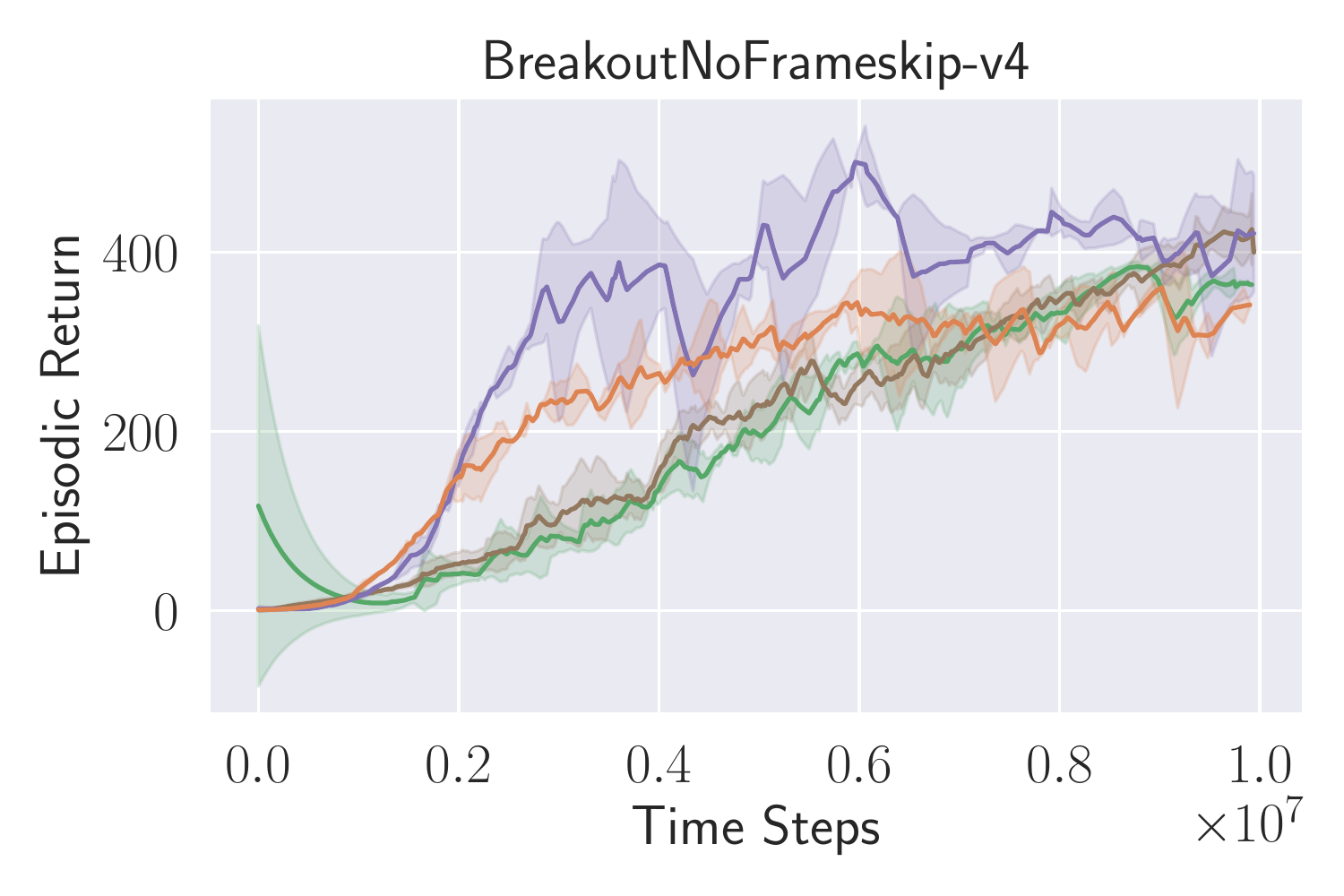}}
{\includegraphics[width=0.32\textwidth]{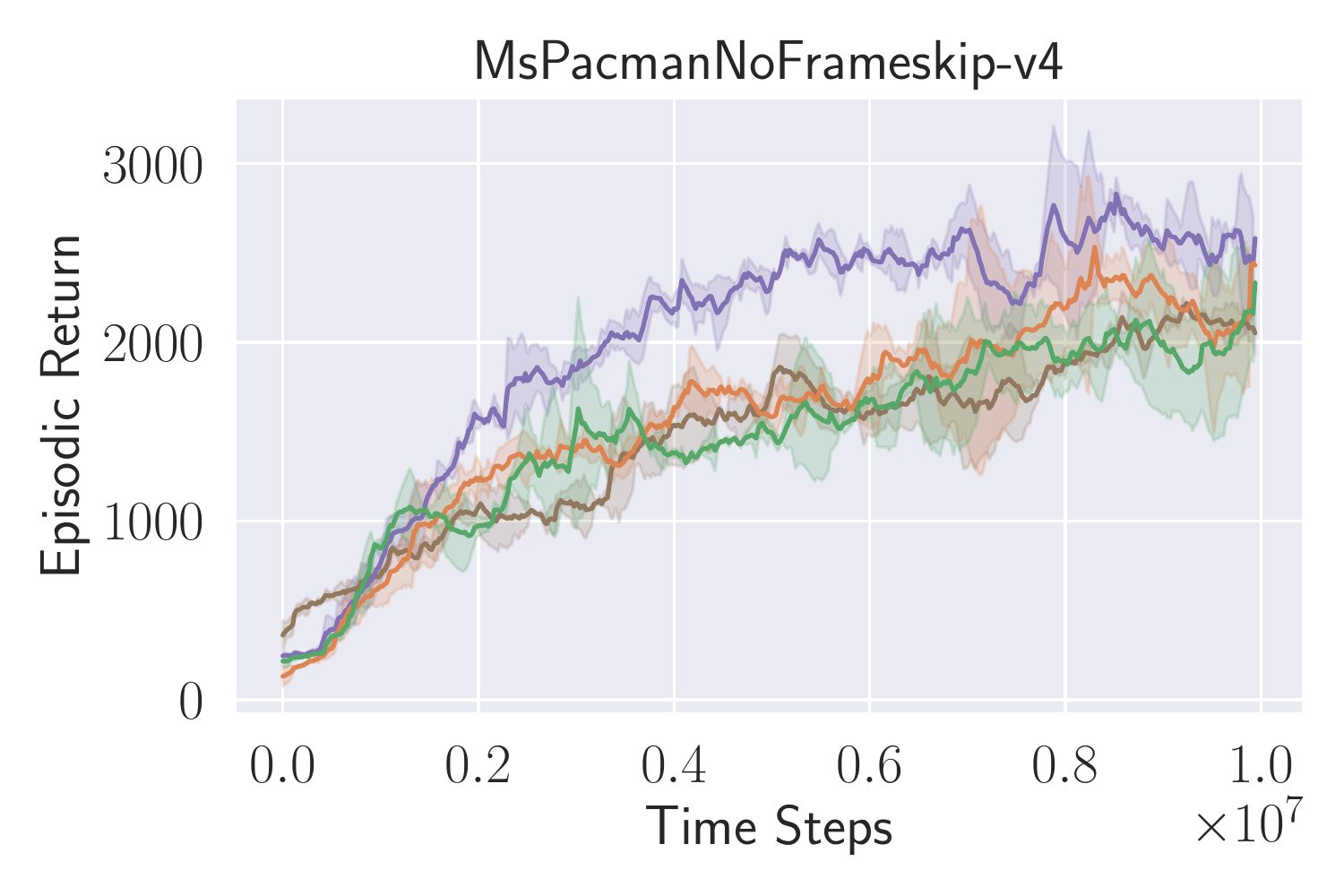}}
{\includegraphics[width=0.32\textwidth]{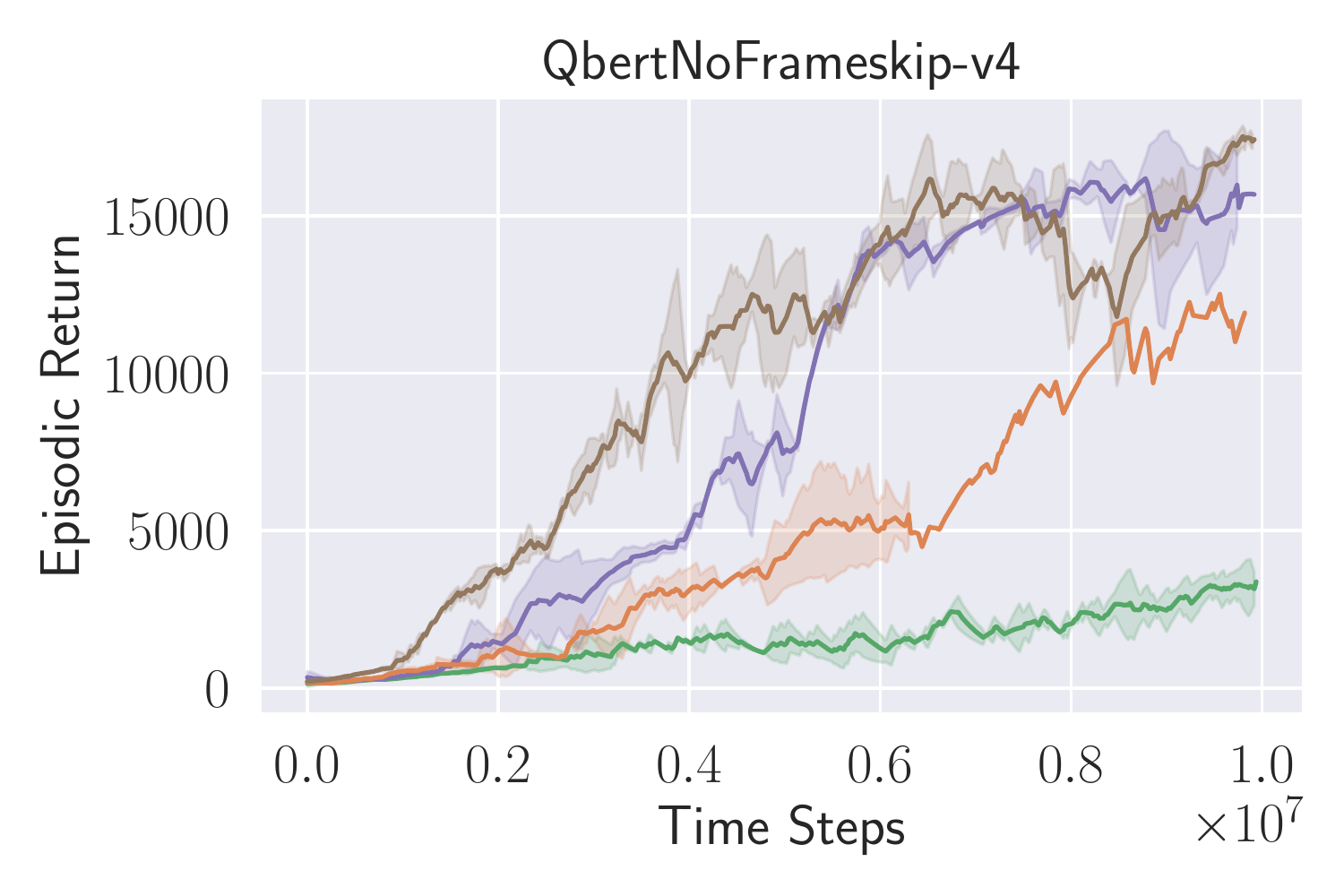}}\hfill
{\includegraphics[width=0.32\textwidth]{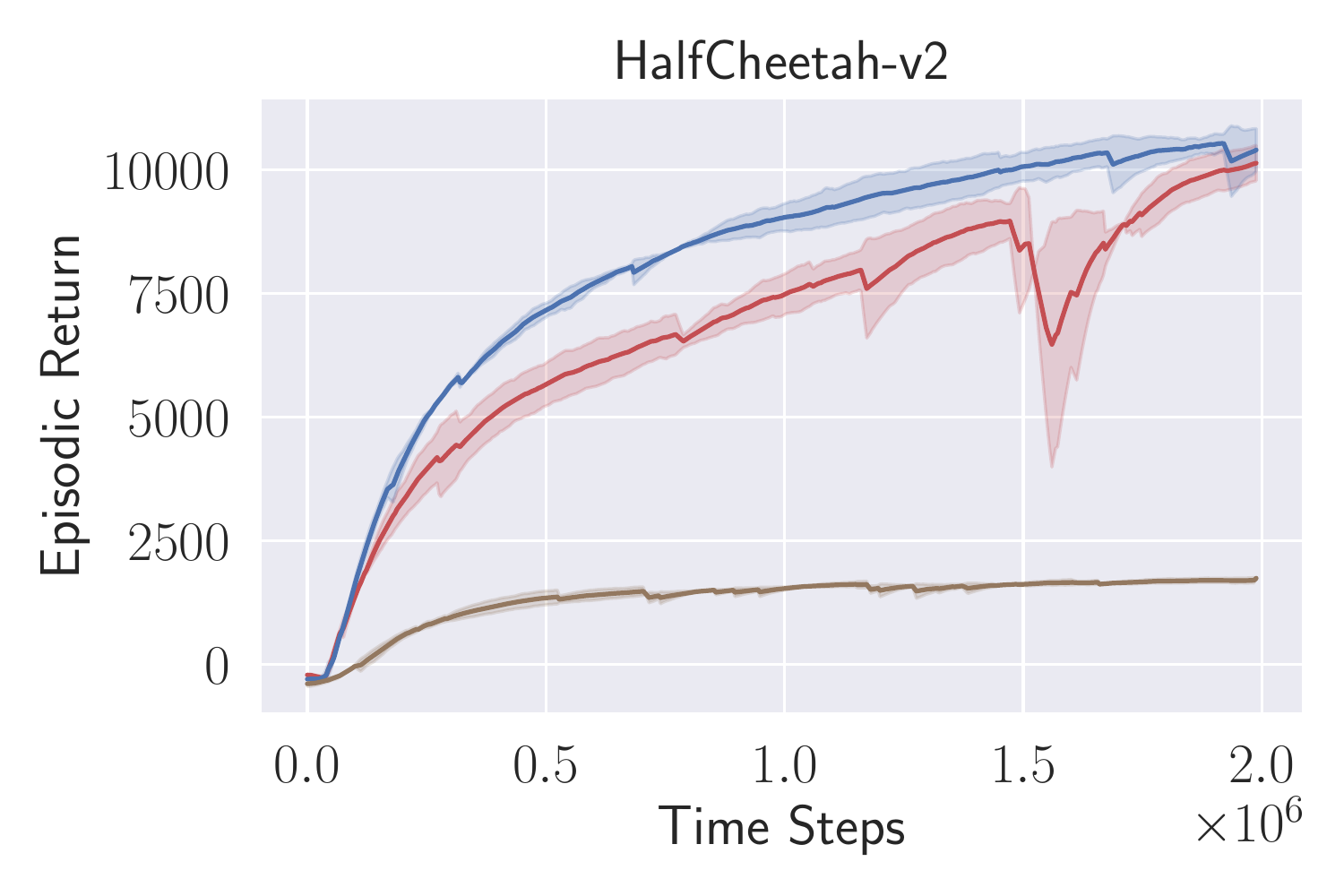}}
{\includegraphics[width=0.32\textwidth]{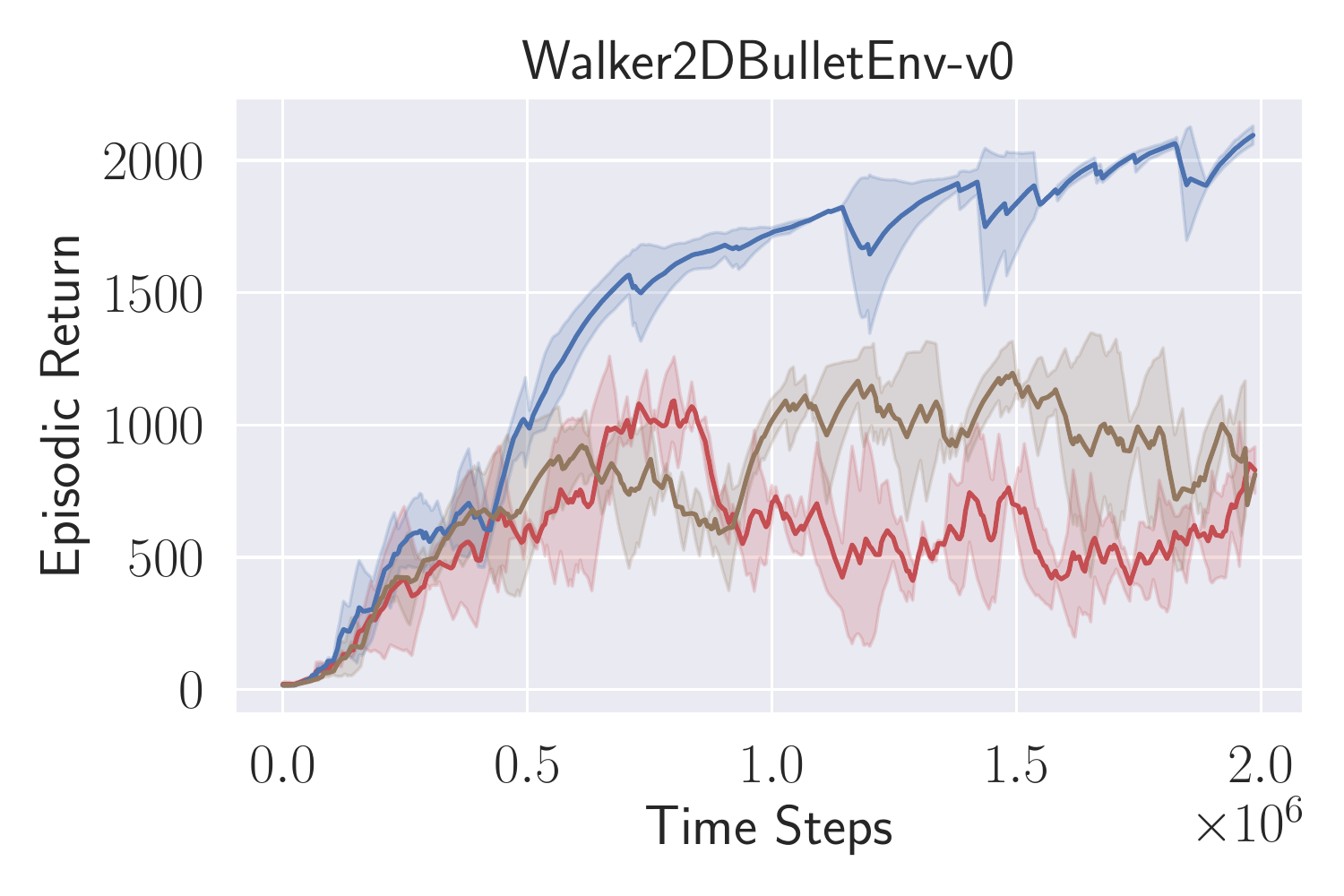}}
{\includegraphics[width=0.32\textwidth]{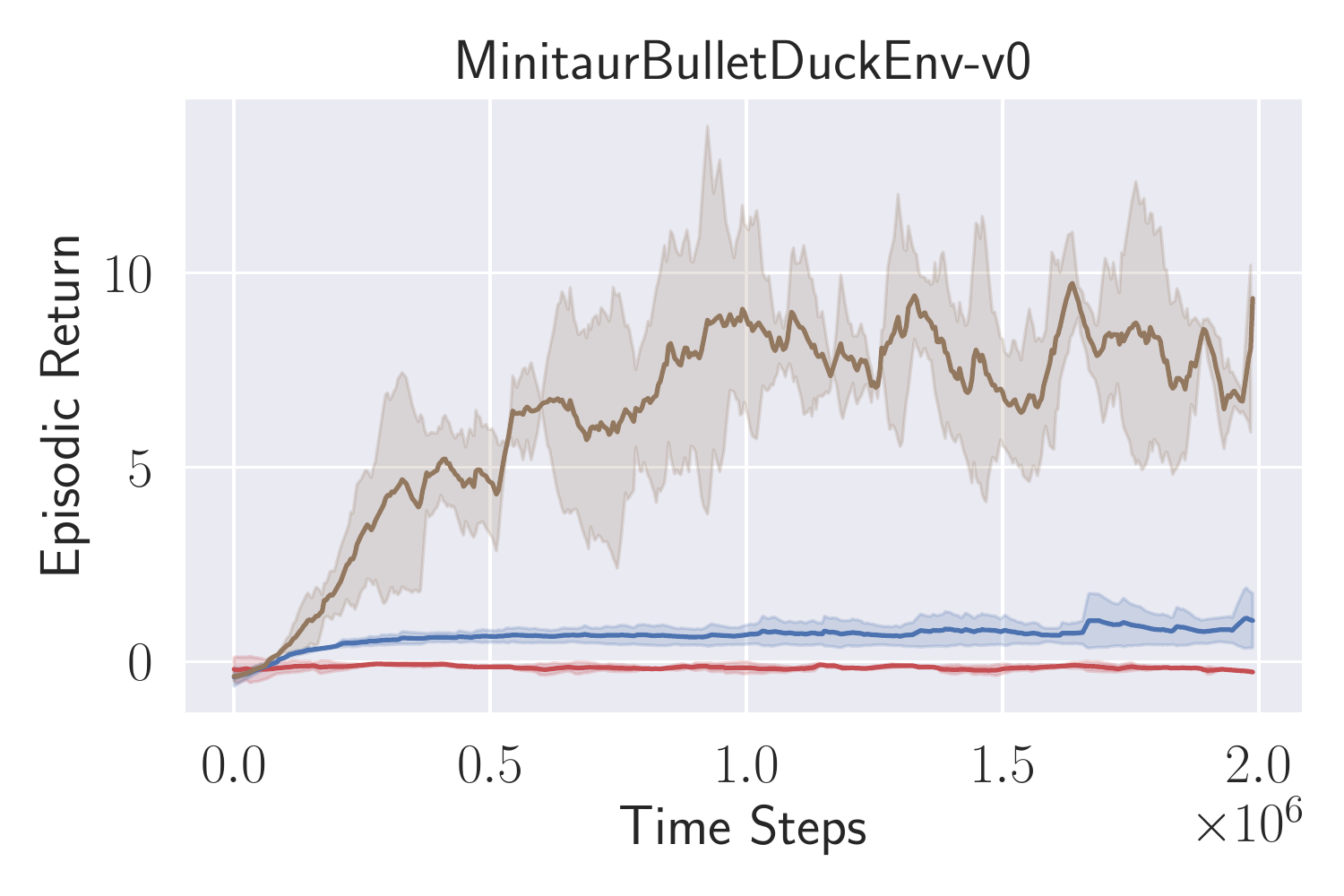}}\hfill
% \subfloat[$4\times4$ Map]{\label{fig:sub:4x4}{\includegraphics[width=0.245\textwidth]{losses_approx_kl/MicrortsMining4x4F9-v0.pdf}}}
% \subfloat[$10\times10$ Map]{\label{fig:sub:10x10}{\includegraphics[width=0.245\textwidth]{losses_approx_kl/MicrortsMining10x10F9-v0.pdf}}}
% \subfloat[$16\times16$ Map]{\label{fig:sub:16x16}{\includegraphics[width=0.245\textwidth]{losses_approx_kl/MicrortsMining16x16F9-v0.pdf}}}
% \subfloat[$24\times24$ Map]{\label{fig:sub:24x24}{\includegraphics[width=0.245\textwidth]{losses_approx_kl/MicrortsMining24x24F9-v0.pdf}}}
  \caption{The performance of \texttt{CleanRL}'s algorithms in Atari and continuous action domains.
%   The shaded area represents one standard deviation of the data over 2 random seeds. Curves are exponentially smoothed with weight of 0.95 for readability.
  }
  \label{fig:benchmark}
\end{figure*}

\section{Cloud Integration}
Despite succinct implementations, we are able to use \texttt{CleanRL} for large-scale studies by leveraging public cloud infrastructure such as Amazon Web Services (AWS). We package the code into a \texttt{Docker} container to orchestrate the experiments on any cloud provider. When needing to run experiments on demand, the users of \texttt{CleanRL} can use \texttt{Terraform} to reproduce a preset infrastructure in a few commands, then utilize our utility scripts to create as many experiments as needed. In 2020 alone, the authors have run over 50,000+ hours of experiments using this workflow. See \url{https://docs.cleanrl.dev} for more documentation.

\section{Open RL Benchmark}
We have created a project called Open RL Benchmark (\url{https://benchmark.cleanrl.dev}) to track canonical RL experiments. To date, we have tracked thousands of experiments across 7+ algorithms in various domains (e.g. Atari, MuJoCo, PyBullet, Procgen~\citep{cobbe2020leveraging}, Griddly~\citep{bamford2020griddly}, Gym-$\mu$RTS~\citep{huang2021gym}). See Figure~\ref{fig:benchmark} for selected experiments. Our benchmark is interactive as shown in Figure~\ref{fig:wandb}: researchers can easily query information such as GPU utilization and videos of an agent's gameplay that are normally hard to acquire in other RL benchmarks.

% and use AWS Batch to run thousands of experiments concurrently. To make the infrastructure easy to manage and reproducible, we use Terraform to spin up services.

\section{Conclusion}
In this paper, we introduced \texttt{CleanRL} as a deep RL library that provides high-quality single-file implementations with experiment tracking and cloud integration. The source code can be found at \url{https://github.com/vwxyzjn/cleanrl}.

% {\noindent \em Remainder omitted in this sample. See http://www.jmlr.org/papers/ for full paper.}

% Acknowledgements should go at the end, before appendices and references
\newpage
% \acks{We would like to acknowledge support for this project
% from the National Science Foundation (NSF grant IIS-9988642)
% and the Multidisciplinary Research Program of the Department
% of Defense (MURI N00014-00-1-0637). }


% 

\vskip 0.2in
\bibliography{sample}

\newpage

\appendix
\section*{Appendix A. Interactive Shell}
\label{app:shell}

% Note: in this sample, the section number is hard-coded in. Following
% proper LaTeX conventions, it should properly be coded as a reference:

%In this appendix we prove the following theorem from
%Section~\ref{sec:textree-generalization}:

% In this appendix we provide supplementary materials.

In \texttt{CleanRL}, we have put most of the variables in the \emph{global python name scope}. This makes it easier to inspect the variables and their shapes. The following figure shows a screenshot of the Spyder editor~\footnote{\url{https://www.spyder-ide.org/}}, where the code is on the left and the interactive shell is on the right. In the interactive shell, we can easily inspect the variables for debugging purposes without modifying the code.

\begin{figure*}[h]
    \centering
    % \makebox[\textwidth][c]{\includegraphics[width=1.3\textwidth]{images/shell2.png}}%
    \includegraphics[width=1\columnwidth]{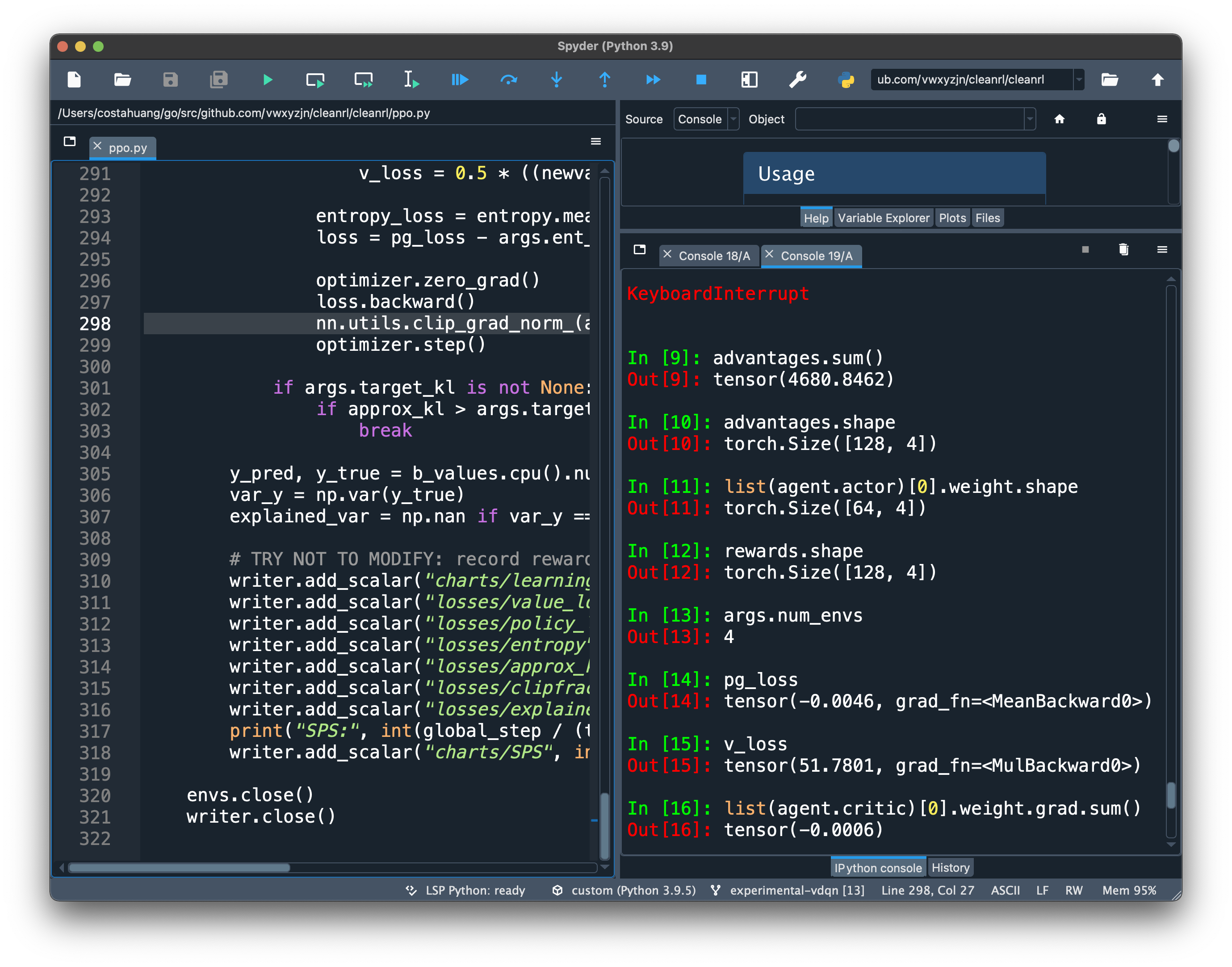}
    % \caption{Using the interactive shell of the Spyder editor to check out variables for debugging purposes.}
    % \label{fig:shell}
\end{figure*}

\newpage
\section*{Appendix B. W\&B Editing Panel}
A screenshot of the W\&B panel that allows the the users to change smoothing weight, add panels to show different metrics like losses, visualize the videos of the agents' gameplay, filter, group, sort, and search for desired experiments.
\begin{figure*}[h]
    \centering
    \makebox[\textwidth][c]{\includegraphics[width=1.2\textwidth]{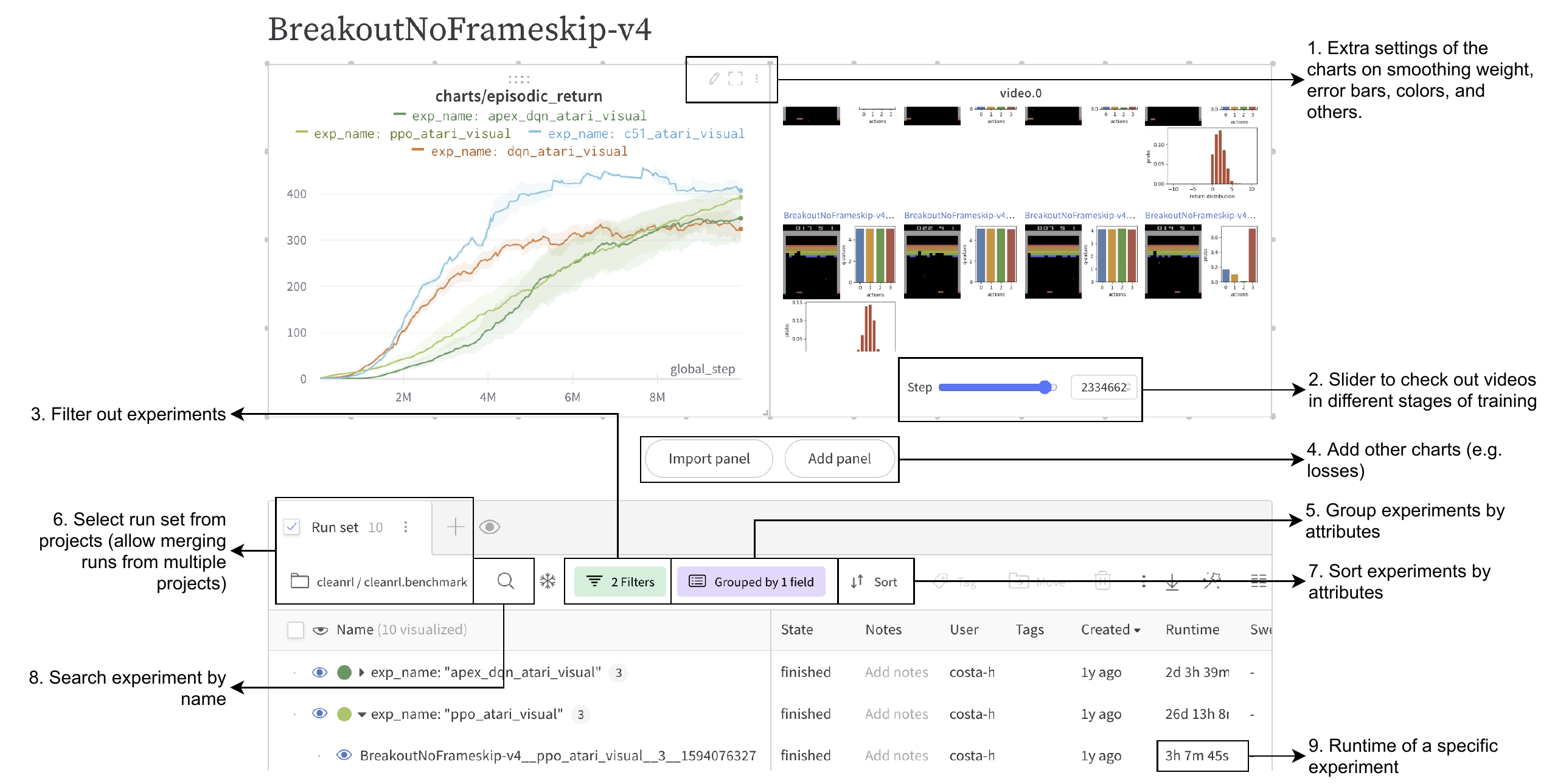}}
\end{figure*}

\end{document}